\relax
\documentclass[letterpaper]{article} % DO NOT CHANGE THIS
\usepackage{aaai22}  % DO NOT CHANGE THIS
\usepackage{times}  % DO NOT CHANGE THIS
\usepackage{helvet}  % DO NOT CHANGE THIS
\usepackage{courier}  % DO NOT CHANGE THIS
\usepackage[hyphens]{url}  % DO NOT CHANGE THIS
\usepackage{graphicx} % DO NOT CHANGE THIS
\usepackage{natbib}  % DO NOT CHANGE THIS AND DO NOT ADD ANY OPTIONS TO IT
\usepackage{caption} % DO NOT CHANGE THIS AND DO NOT ADD ANY OPTIONS TO IT
\DeclareCaptionStyle{ruled}{labelfont=normalfont,labelsep=colon,strut=off} % DO NOT CHANGE THIS
\usepackage{algorithm}
\usepackage{algorithmic}

\usepackage[utf8]{inputenc} % allow utf-8 input
\usepackage{url}            % simple URL typesetting
\usepackage{booktabs}       % professional-quality tables
\usepackage{amsfonts}       % blackboard math symbols
\usepackage{nicefrac}       % compact symbols for 1/2, etc.
\usepackage{microtype}      % microtypography
\usepackage{lipsum}
\usepackage{graphicx}

\usepackage{xcolor}
\usepackage{multirow}
\usepackage{array}
\usepackage{rotating} 
\usepackage{amssymb}
\usepackage{amsmath}
\usepackage{tabularx}
\usepackage[switch]{lineno}
\graphicspath{ {./images/} }

\usepackage{color, colortbl}
\definecolor{Gray}{gray}{0.9}

\usepackage{newfloat}
\usepackage{listings}
\floatstyle{ruled}
\newfloat{listing}{tb}{lst}{}
\floatname{listing}{Listing}
\title{TVT: Transferable Vision Transformer for Unsupervised Domain Adaptation}
\author{
    Jinyu Yang\textsuperscript{\rm 1}, 
    Jingjing Liu\textsuperscript{\rm 2}, 
    Ning Xu\textsuperscript{\rm 2},
    Junzhou Huang\textsuperscript{\rm 1}
}
\affiliations{
    \textsuperscript{\rm 1}University Of Texas at Arlington\\
    \textsuperscript{\rm 2}Kuaishou Techlology
}

\usepackage{bibentry}
\begin{document}
\maketitle

\begin{abstract}
Unsupervised domain adaptation (UDA) aims to transfer the knowledge learnt from a labeled source domain to an unlabeled target domain. Previous work is mainly built upon convolutional neural networks (CNNs) to learn domain-invariant representations. With the recent exponential increase in applying Vision Transformer (ViT) to vision tasks, the capability of ViT in adapting cross-domain knowledge, however, remains unexplored in the literature. To fill this gap, this paper first comprehensively investigates the transferability of ViT on a variety of domain adaptation tasks. Surprisingly, ViT demonstrates superior transferability over its CNNs-based counterparts with a large margin, while the performance can be further improved by incorporating adversarial adaptation. Notwithstanding, directly using CNNs-based adaptation strategies fails to take the advantage of ViT's intrinsic merits (e.g., attention mechanism and sequential image representation) which play an important role in knowledge transfer. To remedy this, we propose an unified framework, namely Transferable Vision Transformer (TVT) \footnote{\href{https://github.com/uta-smile/TVT}{https://github.com/uta-smile/TVT}}, to fully exploit the transferability of ViT for domain adaptation. Specifically, we delicately devise a novel and effective unit, which we term Transferability Adaption Module (TAM). By injecting learned transferabilities into attention blocks, TAM compels ViT focus on both transferable and discriminative features. Besides, we leverage discriminative clustering to enhance feature diversity and separation which are undermined during adversarial domain alignment. To verify its versatility, we perform extensive studies of TVT on four benchmarks and the experimental results demonstrate that TVT attains significant improvements compared to existing state-of-the-art UDA methods. 
\end{abstract}

\section{Introduction}
Deep neural networks (DNNs) demonstrate unprecedented achievements on various machine learning problems and applications.
However, such impressive performance heavily relies on massive amounts of labeled data which requires considerable time and labor efforts to collect.
Therefore, it is desirable to train models that can leverage rich labeled data from a different but related domain and generalize well on target domains with no or limited labeled examples.
Unfortunately, the canonical supervised-learning paradigm suffers from the domain shift issue that poses a major challenge in adapting models across domains.
This motivates the research on unsupervised domain adaptation (UDA) \cite{wang2018deep} which is a special scenario of transfer learning \cite{pan2009survey}.
The key idea of UDA is to project data points of the labeled source domain and the unlabeled target domain into a common feature space, such that the projected features are both discriminative (semantic meaningful) and domain-invariant, in turn, generalize well to bridge the domain gap. 
To achieve this goal, various methods have been proposed in the past decades, among which adversarial adaptation has become the dominant technique in this field, which attempts to align cross-domain representations by minimizing an adversarial loss through a domain discriminator \cite{ganin2016domain,tzeng2017adversarial,long2017conditional}.

Recently, Vision Transformer (ViT) \cite{dosovitskiy2020image} has received increasing attention in the vision community. Different from CNNs that act on local receptive fields of the given image, ViT models long-range dependencies among visual features across the entire image, through the global self-attention mechanism.
Specifically in ViT, each image is split into a sequence of fixed-size non-overlapping patches, which are then linearly embedded and concatenated with position embeddings.
To be consistent with NLP paradigm, a class token is prepended to the patch tokens, serving as the representation of the whole image. 
Then, those sequential embeddings are fed into a stack of transformers to learn desired visual representations. Due to its advantages in global context modeling, ViT has obtained excellent results on various vision tasks, such as image classification~\cite{dosovitskiy2020image}, object detection~\cite{carion2020end,wang2021pyramid}, segmentation \cite{zheng2020rethinking,liu2021swin}, and video understanding \cite{girdhar2019video,neimark2021video}.

%two primary strategies are commonly used in existing UDA methods:
%(i) minimize some measure of domain shift such as maximum mean discrepancy (MMD) \cite{tzeng2014deep,long2015learning}, and

Despite that ViT is becoming increasingly popular, two important questions related to domain adaption remain unanswered.
First, \emph{how transferable is ViT across different domains, compared to its CNNs counterparts?}
As ViT is convolution-free and lacks some inductive bias inherent to CNNs (e.g., locality and translation equivariance), it relies on large-scale pre-training to trump inductive bias.
Such training prerequisite along with the learned global attentions may provide ViT with outstanding capability in domain transferring, yet this hypothesis has not been investigated.
The second question is, \emph{how can we properly improve ViT in adapting different domains?}
One intuitive approach is to directly apply adversarial discriminator onto the class tokens to perform adversarial alignment, where the state of a class token represents the entire image. However, cross-domain alignment of such global features assumes all regions or aspects of the image have the equal transferability and discriminative potential, which is not always tenable. 
For instance, background regions can be easier aligned across domains, while foreground regions are more discriminative.
In other words, some discriminative features may lack transferability, and some transferable features may not contribute much to the downstream task (e.g., classification).
Therefore, in order to properly enhance the transferability of ViT, it is critical to identify fine-grained features that are both transferable and discriminative.
%(ii) simply aligning global representations ignores the complex distributions of different regions.
%This is evident by the recent studies that fine-grained alignment of cross-domain features play an important role in domain adaptation \cite{pei2018multi,wang2019transferable}.
%Although this problem can be alleviated by introducing multiple region-level domain discriminators \cite{wang2019transferable}, this strategy dramatically increases the model complexity and training difficulty.

In this paper we aim to present our answers to the two aforementioned questions. Firstly, to fill the blank of understanding ViT's transferability, we first conduct a comprehensive study of vanilla ViT \cite{dosovitskiy2020image} on public UDA benchmarks.
As expected, our experimental results demonstrate that ViT is more transferable than its strong CNNs-based counterparts, which can be partially explained by the global context modeling and large-scale pre-training.
Besides, we observe further improvements by applying an adversarial discriminator to the class tokens of ViT, which only aligns global representations.
However, such strategy suffers from the oversimplified assumption and ignores the inherent properties of ViT that are beneficial for domain adaptation: i) sequential patch tokens actually give us the free access to fine-grained features; ii) the self-attention mechanism in transformer naturally works as a discriminative probe.
In the light of this, we propose an unified UDA framework that makes full use of ViT's inherent merits. We name it Transferable Vision Transformer (TVT).

The key idea of our method is to retain both transferable and discriminative features which are essential in knowledge adaptation.
To achieve this goal, we first introduce the novel Transferability Adaption Module (TAM) built upon a conventional transformer. 
TAM uses a patch-level domain discriminator to measure the transferabilities of patch tokens, and injects learned transferabilities into the multi-head self-attention block of a transformer. 
On one hand, the attention weights of patch tokens in the self-attention block are used to determine their semantic importance, i.e., the features with larger attention are more discriminative yet without transferability guarantees.
On the other hand, as patch tokens can be regarded as fine-grained representations of an image, the higher transferability of a token means the local features are more transferable across domains though not necessarily discriminative.
By simply replacing the last transformer of ViT with a plug-and-play TAM, we could drive ViT to focus on both transferable and discriminative features. 
% Empirically, we find using one TAM in the very last works fairly good enough. 

Since our method performs adversarial adaptation that forces the learned features of two domains to be similar, one underlying side-effect is that the discriminative information of target domain might be destroyed during feature alignment.
To address this problem, we design a Discriminative Clustering Module (DCM) inspired by the clustering assumption.
The motivation is to enforce the individual target prediction close to one-hot encoding (well separated) and the global target prediction to be uniformly distributed (global diverse), such that the learnt target-domain representation could retain maximum discriminative information about the input values.

Contributions of this paper are summarized as follows:
\begin{itemize}
\item As far as we know, we are the first investigating the capability of ViT in transferring knowledge on the domain adaptation task. We believe this work gives good insights to understand and explore ViT's transferability while applied to various vision tasks.
\item We propose TAM that delicately leverages the intrinsic characteristics of ViT, such that our method can capture both transferable and discriminative features for domain adaptation. Moreover, we adopt discriminative clustering assumption to alleviate the discrimination destruction during adversarial alignment.
\item Without any bells and whistles, our method set up a new competitive baseline cross several public UDA benchmarks.
\end{itemize}

%To match the cross-domain representation, we follow the paradigm of adversarial discriminator.
%In addition to the traditional global alignment, we introduce a local alignment strategy which is guided by an adaptive adversarial loss.
%The key idea is to reduce the side effect of adversarial learning XXX.
%To the best of our knowledge neither the use of attacks to improve SSL nor the design of adversarial examples with this property have been previously discussed in the literature.

\section{Related Work}
\paragraph{Unsupervised Domain Adaptation}
Transfer learning aims to learn transferable knowledge that are generalizable across different domains with different distributions \cite{pan2009survey,ying2018transfer}.
This is built upon the evidence that feature representations in machine learning models, especially in deep neural networks, are transferable \cite{yosinski2014transferable}.
The main challenge of transfer learning is to reduce the domain shift or the discrepancy of the marginal probability distributions across domains \cite{wang2018deep}.
In the past decades, various methods have been proposed to address one canonical transfer learning problem, i.e., unsupervised domain adaptation (UDA), where no labels are available for the target domain.
%The intuition is to learn a representation that minimizes the distance between the source and target distributions.
For instance, DDC \cite{tzeng2014deep} attempted to learn domain-invariant features by minimizing Maximum Mean Discrepancy (MMD) \cite{borgwardt2006integrating} between two domains.
%where MMD represents the standard distribution distance metric.
Long et al. further improved DDC by embedding hidden representations of all task-specific layers in a reproducing Hilbert space and used a multiple kernel variant of MMD to measure the domain distance \cite{long2015learning}.
%Added  However, those methods only encourage the matching of the marginal distributions, and therefore fail to consider the shift in the joint distributions of input features and output labels. (show less)
Long et al. proposed to align joint distributions of multiple domain-specific layers across domains through a joint maximum mean discrepancy metric \cite{long2017deep}.
Another line of effort was inspired by the success of adversarial learning \cite{goodfellow2014generative}. By introducing a domain discriminator and modeling the domain adaption as a minimax problem \cite{ganin2016domain,tzeng2017adversarial,long2017conditional}, an encoder is trained to generate domain-invariant features, through deceiving a discriminator which tries to distinguish features of source domain from that of target domain.

It is noteworthy that all of these methods completely or partially used CNNs as the fundamental block \cite{lecun1998gradient,krizhevsky2012imagenet,he2016deep}. 
By contrast, our method explores ViT \cite{dosovitskiy2020image} to tackle the UDA problem, as we believe ViT has better potential and capability in domain adaptation owning to some of its properties.
Although previous UDA methods (e.g., adversarial learning) are able to improve vanilla ViT to some extent, they were not well designed for transformer-based models, and thereby cannot leverage ViT's inherent characteristic of providing attention information and fine-grained representations.
However, Our method is delicately designed with the nature of ViT and could effectively leverages the transferability and discrimination of each feature for knowledge transfer, thus having better chance in fully exploiting the adaptation power of ViT.
%To this end, we propose XXX to effectively generate transferable attention and align local regions guided by attention.

%\vspace{-10pt}
\paragraph{Vision Transformer}
Transformers \cite{vaswani2017attention} was firstly proposed in the NLP field and demonstrate record-breaking performance on various language
tasks, e.g., text classification and machine translation \cite{devlin2018bert,beltagy2020longformer,zhou2020informer}.
Much of such impressive achievement is attributed to the power of capturing long-range dependencies through attention mechanism.
Spurred by this, some recent studies attempted to integrate attention into CNNs to augment feature maps, aiming to provide the capability in modeling heterogeneous interactions \cite{wang2018non,bello2019attention,hu2018relation}.
Another pioneering work of completely convolution-free architecture is Vision Transformer (ViT), which applied transformers on a sequence of fixed-size non-overlapping image patches.
Different from CNNs that rely on image-specific inductive biases (e.g., locality and translation equivariance), ViT takes the benefits from large-scale pre-training data and global context modeling.
One such method \cite{dosovitskiy2020image}, known for its simplicity and accuracy/compute trade-off, competes favorably against CNNs on the classification task and lays the foundation for applying transformer to different vision tasks.
ViT and its variants have proved their wide applicability in object detection \cite{carion2020end,zhu2020deformable,wang2021pyramid}, segmentation \cite{zheng2020rethinking,wang2020end}, and video understanding \cite{girdhar2019video,neimark2021video}, etc. 
%Some other efforts have been endeavored to reduce the computational complexity of self-attention of ViT, while retaining the global dependency paths cross the entire image (\cite{chen2021regionvit,liu2021swin,chu2021twins}.
% For instance, RegionViT adopts pyramid structure and employs a novel regional-to-local attention rather than global self-attention in ViT \cite{chen2021regionvit}, Swin Transformer proposes to use shifted windows to improve the computational efficiency ,
% Twins uses locally-grouped self-attention and global sub-sampled attention to capture local and global dependencies, respectively,
% NesT further extends ViT in a hierarchical way together with a block aggregation function \cite{zhang2021aggregating},
% ViT-G scales ViT with two billion parameters and reaches a new state of the art \cite{zhai2021scaling}.
 
Despite the success of ViT on different vision tasks, to the best of our knowledge, neither their transferability nor the design of UDA methods with ViT have been previously discussed in the literature.
To this end, we focus in this paper on the investigation of ViT's capability in knowledge transferring across different domains.
We propose a novel UDA framework tailored for ViT by exploring its intrinsic merits and prove its superiority over existing methods.

% In this section, we first briefly recap the preliminaries of adversarial learning UDA for the classification task and the self-attention mechanism of ViT.
% We then detail the proposed UDA framework Transferable Vision Transformer (TVT), which explores adversarial learning for vanilla ViT with two new adaptation modules.

%In what follows, we investigate the transferability of ViT and conduct the early attempts to improve ViT's ability in knowledge transfer by incorporating adversarial learning (ViT+Adv).
%Although it is trival to apply adversarial learning to ViT to improve their transferability, such strategy does not take the intrinsic properties of ViT into account, which may lead to negative transfer.
%Finally, we detail our UDA framework known as transferable vision transformer (TVT).

%\vspace{-10p}
\section{Preliminaries} 
\paragraph{Adversarial Learning UDA}
We consider the image classification task in UDA, where a labeled source domain $\mathcal{D}_s\{(x_i^s, y_i^s)\}^{n_s}_{i=1}$ with $n_s$ examples and an unlabeled target domain $\mathcal{D}_t\{x_j^t\}^{n_t}_{j=1}$ with $n_t$ examples are given.
The goal of UDA is to learn features that are both discriminative and invariant to the domain discrepancy, and in turn guarantee accurate prediction on the unlabeled target data.
Here, a common practice is to jointly performs feature learning, domain adaptation, and classifier learning by optimizing the following loss function:
\begin{equation} \label{eq:1}
\begin{aligned}
%\mathcal{L}_{basic}(\mathcal{X}_s, \mathcal{X}_{t}) = {} &
%\frac{1}{n_s}\sum_{x_i \in \mathcal{D}_s} \mathcal{L}_{ce}(G_c(G_f({x}^s_i)), y^s_i) + 
%\alpha \mathcal{L}_{dis}(G_f({x}^s), G_f({x}^{t})),
\mathcal{L}_{clc}({x}^s, y^s) + \alpha \mathcal{L}_{dis}({x}^s, {x}^t)
\end{aligned}
\end{equation}
%where $G_f$ is an encoder, $G_c$ is a classifier, $ \mathcal{L}_{ce} $ is standard cross-entropy loss, 
where $\mathcal{L}_{clc}$ is supervised classification loss, $\mathcal{L}_{dis}$ is a transfer loss with various possible implementations, and $\alpha$ is used to control the importance of $\mathcal{L}_{dis}$.
One of the most commonly used $\mathcal{L}_{dis}$ is the adversarial loss which encourages a domain-invariant feature space through a domain discriminator \cite{ganin2016domain}.

%\vspace{-10pt}
\paragraph{Self-attention Mechanism}
The main building block of ViT is Multi-head Self-Attention (MSA), which is used in the transformer to capture long-range dependencies \cite{vaswani2017attention}. 
Specifically, MSA concatenates multiple scaled dot-product attention (short for SA) modules, where each SA module takes a set of queries ($\mathbf{Q}$), keys ($\mathbf{K}$), and values ($\mathbf{V}$) as inputs.
%allows parallel computing.
In order to learn dependencies between distinct positions, SA computes the dot products of the query with all keys, and applies a softmax function to obtain the weights on the values.
\begin{equation}
\text{SA} (\mathbf{Q}, \mathbf{K}, \mathbf{V})= \text{softmax}(\frac{\mathbf{Q}\mathbf{K}^T}{\sqrt{d}})\mathbf{V}
\end{equation}
where $d$ is the dimension of $\mathbf{Q}$ and $\mathbf{K}$. With $\text{SA} (\mathbf{Q}, \mathbf{K}, \mathbf{V})$, MSA is defined as:
\begin{equation}
\begin{split}
\text{MSA} (\mathbf{Q}, \mathbf{K}, \mathbf{V}) &= \text{Concat}(\text{head}_1, ..., \text{head}_k) \mathbf{W}^O \\
\text{where}\; \text{head}_i &= \text{SA} (\mathbf{Q}\mathbf{W}_i^Q, \mathbf{K}\mathbf{W}_i^K, \mathbf{V}\mathbf{W}_i^V)
\end{split}
\end{equation}
where $\mathbf{W}_i^Q$, $\mathbf{W}_i^K$, $\mathbf{W}_i^V$ are projections of different heads, $\mathbf{W}^O$ is another mapping function.
Intuitively, using multiple heads allows MSA to jointly attend to information from different representation subspaces at different positions.

\section{Methodology}
In this section, we first investigate ViT's ability in knowledge transfer on various adaptation tasks. 
After that, we conduct the early attempts to improve ViT's transferability by incorporating adversarial learning.
Finally, we introduce our method named Transferable Vision Transformer (TVT), which consists two new adaptation modules to further improve ViT's capability for cross-domain adaptation..

\subsection{ViT's Transferability}
To the best of our knowledge, the transferability of ViT has not been studied in the literature before, although ViT and its variants have shown great success in various vision task.
To probe into ViT's capability of domain adaptation, we choose the vanilla ViT \cite{dosovitskiy2020image} as the backbone in all of our studies, owing to its simplicity and popularity. 
We train vanilla ViT by labeled source data only and assess its transferability by the classification accuracy on target data.
As mentioned above, CNNs-based approaches dominate UDA research in the past decades and demonstrate great successes. 
% Since deep neural network can learn transferable features which generalize well for domain adaptation, CNNs have become the dominant technique in UDA.
Therefore, we compare vanilla ViT with CNNs-based architectures, including LeNet \cite{lecun1998gradient}, AlexNet \cite{krizhevsky2012imagenet}, and ResNet \cite{he2016deep}.
All experiments are performed on well-established benchmarks with standard evaluation protocols.

Take the results on Office-31 dataset for example. As shown in Table~\ref{table:office31}, Source Only ViT obtains impressing classification accuracy 89.45\%, which is much better than its strong CNN opponents AlexNet (70.1\%) and ResNet (76.1\%). Similar phenomenon can be observed in other benchmark results, where ViT competes favorably against, if not better than, the other state-of-the-arts CNNs backbones, as shown in Table~\ref{table:digits},\ref{table:office_home},\ref{table:visda}. Surprisingly, Source Only ViT even outperforms strong CNNs-based UDA approaches without any bells and whistles. For instance, it achieves an average accuracy 78.74\% on Office-Home dataset (Table~\ref{table:office_home}), beating all CNN-based UDA methods. Compared to SHOT \cite{liang2020we} recognized as the best UDA model nowadays, Source Only ViT obtains 7\% absolute accuracy boost, a big step in pushing the frontier of UDA research.
These evidences justify our hypothesis that ViT is more transferable, partially explained by its large-scale pre-training and global context modeling.
However, as observed in Table~\ref{table:digits}, a large gap still exists between the Source Only and Target Only models (88.3\% vs 99.22\%), which indicates further improvement space of ViT's transferability.
%In line with previous CNNs-based UDA methods, we conduct source only and target only on digits datasets with ViT, and perform source only on object datasets.
% As reported in Table~\ref{table:digits}, ViT demonstrates impressing classification accuracy with source only setting.
% Specifically, ViT achieves 88.30\% on average on Digits dataset, which is quite competitive and is much higher than 73.0\% achieved by LeNet.
% Similar phenomenon can also be observed in Table~\ref{table:office31},\ref{table:office_home},\ref{table:visda} where ViT even competes favorably against, if not better than, the other state-of-the-arts CNNs-based methods.
% For instance, ViT achieves an average accuracy 78.74\% over 12 domain pairs in Table~\ref{table:office_home}, relative to 34.3\% and 46.1\% for AlexNet and ResNet, respectively.
% It also boosts approximately 7\% improvement versus 71.8\% for SHOT \cite{liang2020we} which is recognized as the best UDA model nowadays. 

\subsection{ViT w/ Adversarial Adaptation: Baseline}
We first investigate how ViT benefits from adversarial adaptation \cite{ganin2016domain}, which is widely used in CNNs-based UDA methods.
We follow the typical adversarial adaptation fashion that employs an encoder $G_f$ for feature learning, a classifier $G_c$ for classification, and a domain discriminator $D_g$ for global feature alignment.
Here, $G_f$ is implemented as ViT and $D_g$ is applied to output state of the class tokens of the source and target images.
To accomplish domain knowledge adaptation, $G_f$ and $D_g$ play a minimax game: $G_f$ learns domain-invariant features to deceive $D_g$, while $D_g$ distinguishes source-domain features from that of target-domain.
The objective can be formulated as:
\begin{equation}
\begin{split}
\mathcal{L}_{clc}({x}^s, y^s) &= \frac{1}{n_s}\sum_{x_i \in \mathcal{D}_s} \mathcal{L}_{ce}(G_c(G_f({x}^s_i)), y^s_i) \\
\mathcal{L}_{dis}({x}^s, {x}^t) &= - 
\frac{1}{n}\sum_{x_i \in \mathcal{D}} \mathcal{L}_{ce}(D_g(G_f({x}^*_i)), y^d_i),
\end{split}
\end{equation}
where $n=n_s+n_t$, $\mathcal{D} = \mathcal{D}_s \bigcup \mathcal{D}_t$, $\mathcal{L}_{ce}$ is cross-entropy loss, the superscript $*$ can be either $s$ or $t$ to denote a source or a target domain, and $y^d$ denotes the domain label (i.e., $y^d=1$ is source, $y^d=0$ is target).

We denote ViT with adversarial adaptation as our Baseline. As shown in Table~\ref{table:digits},\ref{table:office31},\ref{table:office_home},\ref{table:visda}, Baseline shows 7.8\%, 0.78\%, 1.56\%, and 3.21\% absolute accuracy improvements over vanilla ViT, respectively on the four benchmarks.
Those results reveal that global feature alignment with a domain discriminator helps ViT's transferability.
However, compared with the digit recognition task, Baseline achieves limited improvements on object detection which is more complicated and challenging.
We boils down such observation to a conclusion that simply applying global adversarial alignment cannot exploit ViT's full transferable power, since it fails to consider two key factors: 
(i) not all regions/features are equally transferable or discriminative. For effective knowledge transfer, it is essential to focus on both transferable and discriminative features;
(ii) ViT naturally provides fine-grained features given its forward passing sequential tokens, and attention weights in transformer actually convey discriminative potentials of patch tokens. 
To address these challenges and fully leverage the merits of ViT, a new UDA framework named Transferable Vision Transformer (TVT) is further proposed.
%(ii) complex distributions are involved in local regions, and

\begin{figure}[t]
	\begin{center}
		\includegraphics[width=1\linewidth]{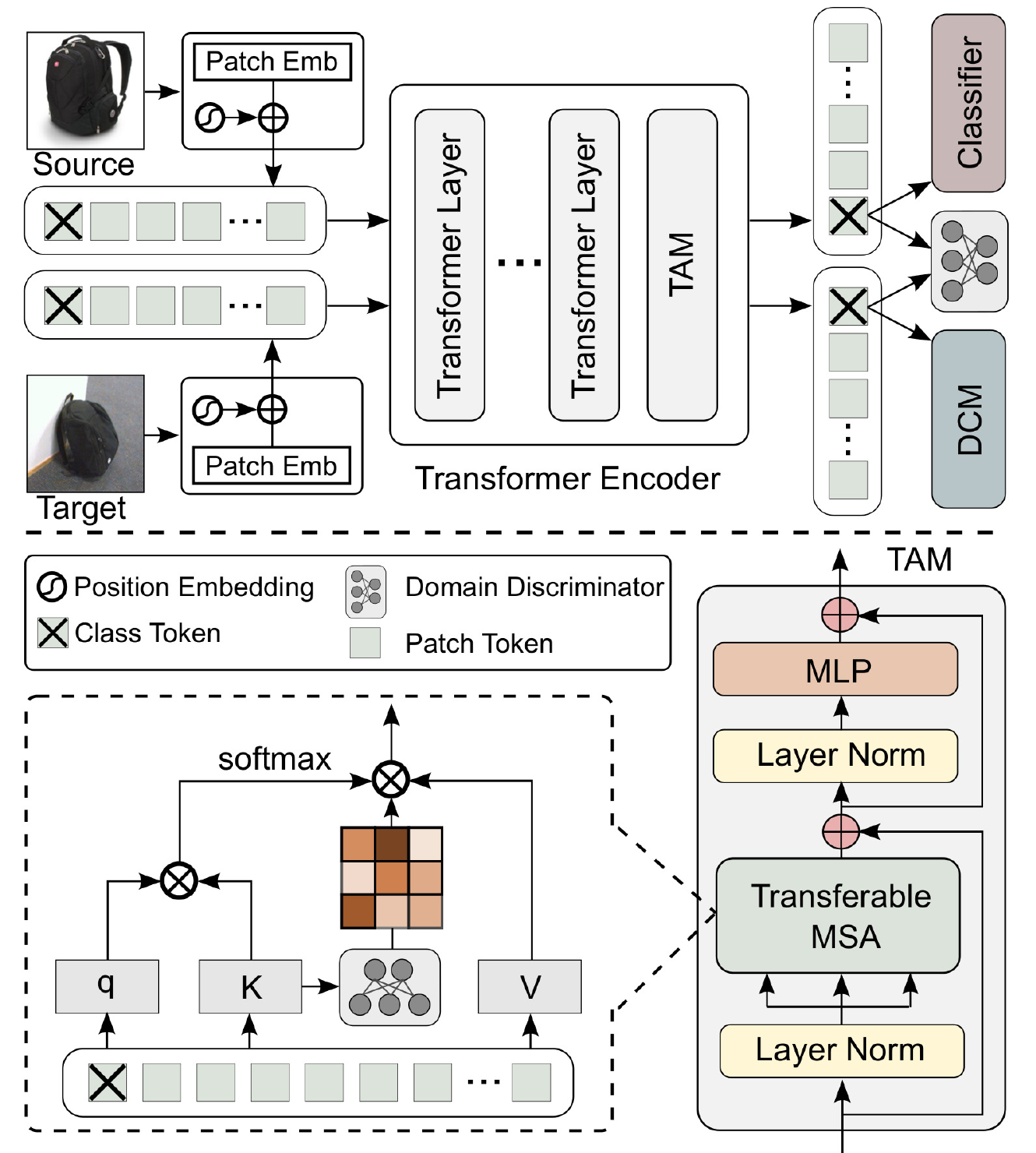}
	\end{center}
	\caption{An overview of the proposed TVT framework. As in ViT, both source and target images are split into fixed-size patches which are then linearly mapped and embedded with positional information. The generated patches are fed into a transformer encoder whose last layer is replaced by Transferability Adaptation Module (TAM). Feature learning, adversarial domain adaptation and classification are accomplished by ViT-akin backbone, two domain discriminators (on patch-level and global-level), Discriminative Clustering Module (DCM) and the MLP-based classifier.}
	\label{fig:framework}
	\vspace{-0.2in}
\end{figure}

\subsection{Transferable Vision Transformer (TVT)}
An overview of TVT is shown in Figure~\ref{fig:framework}, which contains two main modules: 
%(i) a Transferability Adaptation Module (TAM) that explores different transferabilities and discriminative potentials of local features; and
%(ii) a Discriminative Clustering Module (DCM) motivated by the fact that datapoints should be classified with large margin.
(i) a Transferability Adaptation Module (TAM) and (ii) a Discriminative Clustering Module (DCM).
These two modules are highly interrelated and play a complementary role in transferring knowledge for ViT-based architectures. TAM encourages the output state of class token to focus on both transferable and semantic meaningful features, and DCM enforces the aligned features of target-domain samples to be clustered with large margins.
As a consequence, the features learnt by TVT are discriminative in classification and transferable across domains as well.  
We detail each module in what follows.

\subsubsection{Transferability Adaptation Module}
%To mitigate the domain shift in UDA, the common practice is to align cross-domain distributions by making them indistinguishable for a domain discriminator \cite{ganin2016domain}.
As shown in Figure~\ref{fig:framework}, we introduce the Transferability Adaptation Module (TAM) that explicitly considers the intrinsic merits of ViT, i.e., attention mechanisms and sequential patch tokens.
As the patch tokens are regarded as local features of an image, they are corresponded to different image regions or captures different visual aspects as fine-grained representations of an image.
Assuming patch tokens of different semantic importance and transferabilities, TAM aims at assigning different weights to those tokens, to encourage the learned image representations, i.e., the output state of class token, to attend to patch tokens that are both transferable and discriminative. 
%which is referred as transferable attention in what follows.
While the self-attention weights in ViT could be employed as discriminative weights, one major hurdle here is, the transferability of each patch token is not available.
To bypass this difficulty, we adopt a patch-level domain discriminator $D_l$ that matches cross-domain local features \cite{pei2018multi,wang2019transferable} by optimizing:
\begin{equation}
\mathcal{L}_{pat}({x}^s, {x}^t) = - 
\frac{1}{nR}\sum_{x_i \in \mathcal{D}}\sum_{r=1}^R \mathcal{L}_{ce}(D_l(G_f({x}^*_{ir})), y^d_{ir}),
\end{equation}
where $R$ is number of patches, and $D_l(f_{ir})$ is the probability of this region belonging to the source domain.
During adversarial learning, $D_l$ tries to assign $1$ for a source-domain patch and $0$ for the target-domain ones, while $G_f$ combats such circumstances.
Conceptually, a patch that can easily deceive $D_l$ (i.g., $D_l$ is around 0.5) is more transferable across domains and should be given a higher transferability. 
We therefore use $t_{ir}=T(f_{ir})=H(D_l(f_{ir}))\in [0,1]$ to measure the transferability of $r^{th}$ token of $i^{th}$ image, where $H(\cdot)$ is the standard entropy function. 
% We denote the transferability function by $T(\cdot)$.
An other explanation of the transferability is: by assigning weights to different patches, it disentangles an image into common space representations and domain-specific representations, while the passing paths of domain-specific features are softly suppressed.

%Recall that our motivation is to encourage the class token attend to both transferable and discriminative regions.
We then convert the conventional MSA into the transferable MSA (T-MSA) by transferability adaptation, i.e., injecting the learned transferabilities into attention weights of the class token.
Our T-MSA is built upon the transferable self-attention (TSA) block that is formally defined as:
\begin{equation} \label{eq:5}
\text{TSA} (\mathbf{q}, \mathbf{K}, \mathbf{V})= \text{softmax}(\frac{\mathbf{q}\mathbf{K}^T}{\sqrt{d}})\odot [1; T(\mathbf{K}_{patch})] \mathbf{V}
\end{equation}
where $\mathbf{q}$ is the query of the class token, $\mathbf{K}_{patch}$ is the key of the patch tokens, $\odot$ is Hadamard product, and $[;]$ is concatenation operation. 
Obviously, $\text{softmax}(\frac{\mathbf{q}\mathbf{K}^T}{\sqrt{d}})$ and $[1; T(\mathbf{K}_{patch})]$ indicate the discrimination (semantic importance) and the transferability of each patch token, respectively.
To jointly attend to the transferabilities of different representation subspaces and of different locations, we thus define T-MSA as:
\begin{equation}
\begin{split}
\text{T-MSA} (\mathbf{q}, \mathbf{K}, \mathbf{V}) &= \text{Concat}(\text{head}_1, ..., \text{head}_k) \mathbf{W}^O \\
\text{where}\; \text{head}_i &= \text{TSA} (\mathbf{q}\mathbf{W}_i^q, \mathbf{K}\mathbf{W}_i^K, \mathbf{V}\mathbf{W}_i^V)
\end{split}
\end{equation}
Taken them together, we get the TAM as follows:
\begin{equation}
\begin{split}
\hat{\mathbf{z}}^l &= \text{T-MSA}(\text{LN}(\mathbf{z}^{l-1})) + \mathbf{z}^{l-1} \\
\mathbf{z}^l &= \text{MLP}(\text{LN}(\hat{\mathbf{z}}^l)) + \hat{\mathbf{z}}^l
\end{split}
\end{equation}
We only apply TAM to the last transformer layer where patch features are spatially non-local and of higher semantic meanings. By this means, TAM focuses on fine-grained features that are transferable across domains and are discriminative for classification. So we have $l=L$, where $L$ is the total number of transformer layers in ViT. 

%These two operations are implemented by a local discriminator $D_l$ and a transferable MSA (T-MSA), and are highly interrelated, i.e., $D_l$ needs the feedback from   

\subsubsection{Discriminative Clustering Module}
Towards the challenging problem of learning a probabilistic discriminative classifier with unlabeled target data, it is desirable to minimize the expected classification error on the target domain.
However, cross-domain feature alignment through TAM by forcing the two domains to be similar may destroy the discriminative information of the learned representation, if no semantic constrains of the target domain is introduced.
As shown in Figure~\ref{fig:tsne}, although the target feature is indistinguishable from the source feature, it is distributed in a mess which limits its discriminative power.
To address this limitation, we are inspired by the assumptions that: (i) $p^t=\text{softmax}(G_c(G_f(x^t)))$ are expected to retain as much information about $x^t$ as possible \cite{bridle1992unsupervised}; and
(ii) decision boundary should not cross high density regions, but instead lie in low density regions, which is also known as cluster assumption \cite{chapelle2005semi}.
Fortunately, these two assumptions can be met by maximizing mutual information between the empirical distribution on the target inputs and the induced target label distribution \cite{gomes2010discriminative,shi2012information,hu2017learning}, which can be formally defined as:
\begin{equation}
\begin{split}
\mathcal{I}(p^t; x^t) &= H(\bar{p^t}) - \frac{1}{n_t}\sum_{j=1}^{n_t} H(p^t_j) \\
&= - \sum_{k=1}^{K} \bar{p^t_k} log(\bar{p^t_k}) + \frac{1}{n_t}\sum_{j=1}^{n_t}\sum_{k=1}^{K} {p^t_{jk}} log(p^t_{jk})
\end{split}
\end{equation}
where $p^t_j=\text{softmax}(G_c(G_f({x}^t_j)))$, $\bar{p^t}=\mathbb{E}_{x_t}[p^t]$, and $K$ is the number of classes.
Note that maximizing $-\frac{1}{n_t}\sum_{j=1}^{n_t} H(p^t_j)$ enforces the target predictions close to one-hot encoding, therefore the cluster assumption is guaranteed. 
To ensure the global diversity, we also maximize $H(\bar{p^t})$ to avoid that every target data is assigned to the same class.
With $\mathcal{I}(p^t; x^t)$, our model is encouraged to learn tightly clustered target features with uniform distribution, such that the discriminative information in the target domain are retained.

To summarize, the objective function of TVT is:
\begin{equation} \label{eq:10}
\begin{aligned}
\mathcal{L}_{clc}({x}^s, y^s) + \alpha \mathcal{L}_{dis}({x}^s, {x}^t) + \beta \mathcal{L}_{pat}({x}^s, {x}^t) - \gamma \mathcal{I}(p^t; x^t)
\end{aligned}
\end{equation}
where $\alpha$, $\beta$, and $\gamma$ are hyper-parameters.
%The first is that the discriminative model’s decision boundaries should not be located in regions of the input space that are densely populated with datapoints
%maximizing domain similarity that makes the source and the target domains look alike

\section{Experiments}
To verify the effectiveness of our model, we conduct comprehensive studies on commonly used benchmarks and present experimental comparisons against state-of-the-art UDA methods as shown below.
%\vspace{-10pt}
\subsubsection{Digits} is an UDA benchmark on digit classification.
We follow the same setting in previous work to perform adaptations on MNIST \cite{lecun1998gradient}, USPS, and Street View House Numbers (SVHN) \cite{netzer2011reading}. 
For each source-target domain pair, we train our model using the training sets of each domain, and perform evaluations on the standard test set of the target domain.

\paragraph{Office-31} \cite{saenko2010adapting} contains 4,652 images of 31 categories, which were collected from three domains: Amazon (A), DSLR (D), and Webcam (W).
The Amazon (A) image were downloaded from amazon.zom, while the DSLR (D), and Webcam (W) were photoed under the office environment by web and digital SLR camera, respectively. 

%\vspace{-10pt}
\paragraph{Office-Home} \cite{venkateswara2017deep} consists of images from four different domains: Artistic images (Ar), Clip Art (Cl), Product im- ages (Pr), and Real-World images (Rw).
A total of 65 categories are covered within each domain.

%\vspace{-10pt}
\paragraph{VisDA-2017} \cite{peng2017visda} is a synthesis-to-real object recognition task used for the 2018 VisDA challenge. It covers 12 categories.
The source domain contains 152,397 synthetic 2D renderings generated from different angles and under different lighting conditions, while the target domain contains 55,388 real-world images.

%\vspace{-10pt}
\paragraph{Baseline Methods}
% LeNet-based \cite{lecun1998gradient}, AlexNet-based \cite{krizhevsky2012imagenet}, ResNet-based \cite{he2016deep}
We compare with RevGrad \cite{ganin2015unsupervised,ganin2016domain},
ADDA \cite{tzeng2017adversarial}, SHOT \cite{liang2020we}, CDAN \cite{long2017conditional}, CyCADA \cite{hoffman2018cycada}, MCD \cite{saito2018maximum}, DDC \cite{tzeng2014deep}, DAN \cite{long2015learning}, JAN \cite{long2017deep}, PFAN \cite{chen2019progressive}, TADA \cite{wang2019transferable}, ALDA \cite{chen2020adversarial}, TAT \cite{liu2019transferable}, and DTA \cite{lee2019drop}, under the close-set setting where the source and the target domain share the same label space. 
We use the results in their original papers for fair comparison.
For each type of backbone, we report its lower bound performance, denoted as Source Only, meaning the models are trained with source data only.
For digit recognition, we also show the Target Only results as the high-end performance, which is obtained by both training and testing on the labeled target data. Baseline denotes vanilla ViT with adversarial adaptation~\cite{ganin2016domain}.

%\subsubsection{Implementation Details}
% \vspace{-10pt}
% \paragraph{Network Architecture} 
\paragraph{Implementation Details}
The ViT-Base with 16$\times$16 input patch size (or ViT-B/16) \cite{dosovitskiy2020image} pre-trained on ImageNet \cite{deng2009imagenet} is used as our backbone.
The transformer encoder of ViT-B/16 contains 12 transformer layers in total.
%For the baseline, we apply the domain-adversarial discriminator \cite{ganin2016domain} to the source and target class tokens that are generated by the last layer of ViT.
%\vspace{-10pt}
% \paragraph{Experimental Settings}
We train all ViT-based models using mini-batch Stochastic Gradient Descent (SGD) optimizer with the momentum of 0.9.
We initialized the learning rate as 0 and linearly increase it to $lr=0.03$ after 500 training steps. 
We then decrease it by the cosine decay strategy.
The only exception is that we set $lr=0.003$ for D$\rightarrow$ A and W$\rightarrow$ A in Office-31 dataset.
% Hyper-parameters $\alpha$, $\beta$, and $\gamma$ in Equation \ref{eq:10} are chosen from $\{0.1,0.5,1.0\}$, $\{0.01,0.05,0.1\}$, and $\{0.01,0.1,1.0\}$, respectively.

\newcolumntype{g}{>{\columncolor{Gray}[\tabcolsep][0pt]}c}
\begin{table}[t]
	\footnotesize
	\setlength\tabcolsep{9pt}
	\begin{center}
		\begin{tabular}{ @{} l|c|*{3}{c}|g @{} }
			\toprule
			%\multicolumn{24}{ c }{\bf GTA5$\,\to\,$Cityscapes } \\
			%\midrule
			\bf Algorithm & & 
			S→M & U→M & M→U & Avg \\
			\midrule
			
			Source Only &  
			\multirow{7}{*}{\rotatebox[origin=c]{90}{LeNet}} &
			67.1 & 69.6 & 82.2 & 73.0 \\
			
			RevGrad & &
			73.9 & 73.0 & 77.1 & 74.7 \\
			
			ADDA & & 
			76.0 & 90.1 & 89.4 & 85.2 \\
			
			SHOT-IM & &
			89.6 & 96.8 & 91.9 & 92.8 \\
			
			CyCADA  & & 
			90.4 & 96.5 & 95.6 & 94.2 \\
			
			CDAN & &
			89.2 & 98.0 & 95.6 & 94.3 \\
			
			MCD & &
			96.2 & 94.1 & 94.2 & 94.8 \\
			
			\midrule
			Target Only & &
			99.4 & 99.4 & 98.0 & 98.9 \\
			\midrule
			\midrule
			
			Source Only &  
			\multirow{3}{*}{\rotatebox[origin=c]{90}{ViT}} &
			88.58 & 88.23 & 73.09 & 88.30 \\
			
			%{ViT+MMD} & & 
			%92.82 & 98.07 & 96.51 & 95.80 \\
			
			{Baseline} & & 
			92.70 & 98.60 & 97.01 & 96.10 \\
			
			\textbf{TVT} & & 
			99.01 & 99.38 & 98.21 & 98.87 \\
			
			\midrule
			Target Only & & 
			99.70 & 99.70 & 98.26 & 99.22 \\
			\bottomrule
		\end{tabular}
	\end{center}
	\caption{Performance comparison on Digits dataset.}
	\label{table:digits}
 	\vspace{-0.2in}
\end{table}

\begin{table}[t]
	\footnotesize
	\setlength\tabcolsep{0.9pt}
	\begin{center}
		\begin{tabular}{ @{} l|c|*{6}{c}|g @{} }
			\toprule
			%\multicolumn{24}{ c }{\bf GTA5$\,\to\,$Cityscapes } \\
			%\midrule
			\bf Algorithm & & A$\rightarrow$ W & D$\rightarrow$ W & W$\rightarrow$ D & A$\rightarrow$ D & D$\rightarrow$ A & W$\rightarrow$ A & Avg \\
			\midrule
			
			Source Only  &
			\multirow{7}{*}{\rotatebox[origin=c]{90}{AlexNet}} &
			61.6 & 95.4 & 99.0 & 63.8 & 51.1 & 49.8 & 70.1 \\
			
			DDC & &
			61.8 & 95.0 & 98.5 & 64.4 & 52.1 & 52.2 & 70.6 \\
			
			DAN & &
			68.5 & 96.0 & 99.0 & 67.0 & 54.0 & 53.1 & 72.9 \\
			
			RevGrad & & 
			73.0 & 96.4 & 99.2 & 72.3 & 53.4 & 51.2 & 74.3 \\
			
			JAN & & 
			75.2 & 96.6 & 99.6 & 72.8 & 57.5 & 56.3 & 76.3 \\
			
			CDAN & &
			78.3 & 97.2 & 100.0 & 76.3 & 57.3 & 57.3 & 77.7 \\
			
			PFAN & &
			83.0 & 99.0 & 99.9 & 76.3 & 63.3 & 60.8 & 80.4 \\
			
			\midrule
			
			Source Only &  
			\multirow{10}{*}{\rotatebox[origin=c]{90}{ResNet}} &
			68.4 & 96.7 & 99.3 & 68.9 & 62.5 & 60.7 & 76.1 \\
			
			DDC & & 
			75.6 & 96.0 & 98.2 & 76.5 & 62.2 & 61.5 & 78.3 \\
			 
			DAN & & 
			80.5 & 97.1 & 99.6 & 78.6 & 63.6 & 62.8 & 80.4 \\
			 
			RevGrad & & 
			82.0 & 96.9 & 99.1 & 79.7 & 68.2 & 67.4 & 82.2 \\
			 
			JAN & & 
			86.0 & 96.7 & 99.7 & 85.1 & 69.2 & 70.7 & 84.6 \\
			
			CDAN & &
			94.1 & 98.6 & 100.0 & 92.9 & 71.0 & 69.3 & 87.7 \\
			
			TADA & &
			94.3 & 98.7 & 99.8 & 91.6 & 72.9 & 73.0 & 88.4 \\
			
			TAT & &
			92.5 & 99.3 & 100.0 & 93.2 & 73.1 & 72.1 & 88.4 \\
			
			SHOT & &
			90.1 & 98.4 & 99.9 & 94.0 & 74.7 & 74.3 & 88.6 \\
			
			ALDA & &
			95.6 & 97.7 & 100.0 & 94.0 & 72.2 & 72.5 & 88.7 \\
			\midrule
			
			Source Only &  
			\multirow{3}{*}{\rotatebox[origin=c]{90}{ViT}} &
			89.18 & 98.87 & 100.0 & 88.76 & 80.09 & 79.77 & 89.45 \\
			 
			{Baseline} & & 
			91.57 & 98.99 & 100.0 & 90.56 & 80.16 & 80.12 & 90.23 \\
			 
			\textbf{TVT} & & 
			96.35 & 99.37 & 100.0 & 96.39 & 84.91 & 86.05 & 93.85 \\
			\bottomrule
		\end{tabular}
	\end{center}
	\caption{Performance comparison on Office-31 dataset.}
	\label{table:office31}
 	\vspace{-0.2in}
\end{table}

\begin{table*}[t]
	\footnotesize
	\setlength\tabcolsep{1.2pt}
	\begin{center}
		\begin{tabular}{ @{} l|c|*{12}{c}|g @{} }
			\toprule
			%\multicolumn{24}{ c }{\bf GTA5$\,\to\,$Cityscapes } \\
			%\midrule
			\bf Algorithm & & 
			Ar→Cl & Ar→Pr & Ar→Rw & Cl→Ar & Cl→Pr & Cl→Rw & Pr→Ar & Pr→Cl & Pr→Rw & Rw→Ar & Rw→Cl & Rw→Pr & Avg \\
			\midrule
			
			Source Only &  
			\multirow{4}{*}{\rotatebox[origin=c]{90}{AlexNet}} &
			26.4 & 32.6 & 41.3 & 22.1 & 41.7 & 42.1 & 20.5 & 20.3 & 51.1 & 31.0 & 27.9 & 54.9 & 34.3 \\
			
			DAN & &
			31.7 & 43.2 & 55.1 & 33.8 & 48.6 & 50.8 & 30.1 & 35.1 & 57.7 & 44.6 & 39.3 & 63.7 & 44.5 \\
			
			RevGrad & & 
			36.4 & 45.2 & 54.7 & 35.2 & 51.8 & 55.1 & 31.6 & 39.7 & 59.3 & 45.7 & 46.4 & 65.9 & 47.3 \\
			
			JAN & &
			35.5 & 46.1 & 57.7 & 36.4 & 53.3 & 54.5 & 33.4 & 40.3 & 60.1 & 45.9 & 47.4 & 67.9 & 48.2 \\
			
			\midrule
			
			Source Only &  
			\multirow{9}{*}{\rotatebox[origin=c]{90}{ResNet}} &
			34.9 & 50.0 & 58.0 & 37.4 & 41.9 & 46.2 & 38.5 & 31.2 & 60.4 & 53.9 & 41.2 & 59.9 & 46.1 \\
			 
			DAN & & 
			43.6 & 57.0 & 67.9 & 45.8 & 56.5 & 60.4 & 44.0 & 43.6 & 67.7 & 63.1 & 51.5 & 74.3 & 56.3 \\
			 
			RevGrad & & 
			45.6 & 59.3 & 70.1 & 47.0 & 58.5 & 60.9 & 46.1 & 43.7 & 68.5 & 63.2 & 51.8 & 76.8 & 57.6 \\
			 
			JAN & & 
			45.9 & 61.2 & 68.9 & 50.4 & 59.7 & 61.0 & 45.8 & 43.4 & 70.3 & 63.9 & 52.4 & 76.8 & 58.3 \\
			
			CDAN & &
			50.7 & 70.6 & 76.0 & 57.6 & 70.0 & 70.0 & 57.4 & 50.9 & 77.3 & 70.9 & 56.7 & 81.6 & 65.8 \\
			
			TAT & &
			51.6 & 69.5 & 75.4 & 59.4 & 69.5 & 68.6 & 59.5 & 50.5 & 76.8 & 70.9 & 56.6 & 81.6 & 65.8 \\
			
			ALDA & &
			53.7 & 70.1 & 76.4 & 60.2 & 72.6 & 71.5 & 56.8 & 51.9 & 77.1 & 70.2 & 56.3 & 82.1 & 66.6 \\
			
			TADA & &
			53.1 & 72.3 & 77.2 & 59.1 & 71.2 & 72.1 & 59.7 & 53.1 & 78.4 & 72.4 & 60.0 & 82.9 & 67.6 \\
			
			SHOT & & 
			57.1 & 78.1 & 81.5 & 68.0 & 78.2 & 78.1 & 67.4 & 54.9 & 82.2 & 73.3 & 58.8 & 84.3 & 71.8 \\
			\midrule

			Source Only &  
			\multirow{3}{*}{\rotatebox[origin=c]{90}{ViT}} &
			66.16 & 84.28 & 86.64 & 77.92 & 83.28 & 84.32 & 75.98 & 62.73 & 88.66 & 80.10 & 66.19 & 88.65 & 78.74 \\
			
			%ViT+MMD & & 
			%71.23 & 83.74 & 86.21 & 77.96 & 83.42 & 84.26 & 77.96 & 71.11 & 88.73 & 82.08 & 72.60 & 88.83 & 80.68 \\
			
			Baseline & & 
			71.94 & 80.67 & 86.67 & 79.93 & 80.38 & 83.52 & 76.89 & 70.93 & 88.27 & 83.02 & 72.91 & 88.44 & 80.30 \\
			 
			\textbf{TVT} & & 
			74.89 & 86.82 & 89.47 & 82.78 & 87.95 & 88.27 & 79.81 & 71.94 & 90.13 & 85.46 & 74.62 & 90.56 & 83.56 \\
			 
			\bottomrule
		\end{tabular}
	\end{center}
	\caption{Performance comparison on Office-Home dataset.}
	\label{table:office_home}
% 	\vspace{-0.2in}
\end{table*}

\begin{table*}
	\footnotesize
	\setlength\tabcolsep{2.5pt}
	\begin{center}
		\begin{tabular}{ @{} l|c|*{12}{c}|g @{} }
			\toprule
			%\multicolumn{24}{ c }{\bf GTA5$\,\to\,$Cityscapes } \\
			%\midrule
			\bf Algorithm & & 
			plane & bcycl & bus & car & house & knife & mcycl & person & plant & sktbrd & train & truck & Avg \\
			\midrule
			
			Source Only &  
			\multirow{6}{*}{\rotatebox[origin=c]{90}{ResNet}} &
			55.1 & 53.3 & 61.9 & 59.1 & 80.6 & 17.9 & 79.7 & 31.2 & 81.0 & 26.5 & 73.5 & 8.5 & 52.4 \\
			
			RevGrad & &
			81.9 & 77.7 & 82.8 & 44.3 & 81.2 & 29.5 & 65.1 & 28.6 & 51.9 & 54.6 & 82.8 & 7.8 & 57.4 \\
			
			MCD & &
			87.0 & 60.9 & 83.7 & 64.0 & 88.9 & 79.6 & 84.7 & 76.9 & 88.6 & 40.3 & 83.0 & 25.8 & 71.9 \\
			
			ALDA & &
			93.8 & 74.1 & 82.4 & 69.4 & 90.6 & 87.2 & 89.0 & 67.6 & 93.4 & 76.1 & 87.7 & 22.2 & 77.8 \\
			
			DTA & &
			93.7 & 82.2 & 85.6 & 83.8 & 93.0 & 81.0 & 90.7 & 82.1 & 95.1 & 78.1 & 86.4 & 32.1 & 81.5 \\
			
			SHOT & &
			94.3 & 88.5 & 80.1 & 57.3 & 93.1 & 94.9 & 80.7 & 80.3 & 91.5 & 89.1 & 86.3 & 58.2 & 82.9 \\
			\midrule
			
			Source Only &  
			\multirow{3}{*}{\rotatebox[origin=c]{90}{ViT}} &
			98.16 & 72.98 & 82.52 & 62.00 & 97.34 & 63.52 & 96.46 & 29.80 & 68.74 & 86.72 & 96.74 & 23.65 & 73.22 \\
			 
			{Baseline} & & 
			94.60 & 81.55 & 81.81 & 69.85 & 93.54 & 69.93 & 88.60 & 50.45 & 86.79 & 88.47 & 91.45 & 20.10 & 76.43 \\
			 
			%{ViT+Adv+IM} & &
			%95.01 & 89.01 & 78.91 & 61.99 & 93.29 & 87.13 & 85.44 & 78.77 & 94.11 & 93.03 & 90.75 & 48.13 & 82.96 \\
			 
			\textbf{TVT} & & 
			92.92 & 85.58 & 77.51 & 60.48 & 93.60 & 98.17 & 89.35 & 76.40 & 93.56 & 92.02 & 91.69 & 55.73 & 83.92 \\
			\bottomrule
		\end{tabular}
	\end{center}
	\caption{Performance comparison on VisDA-2017 dataset.}
	\label{table:visda}
	\vspace{-0.2in}
\end{table*}

%\vspace{-10pt}
\subsubsection{Results of Digit Recognition}
For the digit recognition task, we perform evaluations on SVHN$\rightarrow$MNISt, USPS$\rightarrow$MNIST, and MNIST$\rightarrow$USPS, following the standard evaluation protocol of UDA.
Shown in Table~\ref{table:digits}, TVT obtains the best mean accuracy for each task and outperforms prior work in terms of the average classification accuracy. TVT also performs better than Baseline (+2.7\%) due to the contribution of the proposed TAM and DCM. 
In particular, TVT achieves comparable results to Target Only model, indicating that the domain shift problem is well alleviated. 
% In particular, we outperform both source only and baseline, owing to the ability of our model in (i) capturing both transferable and discriminative features and (ii) retaining discriminative information while searching for the domain-invariant representations.

%\vspace{-10pt}
\subsubsection{Results of Object Recognition}
\begin{figure}[t]
	\begin{center}
		\includegraphics[width=1\linewidth]{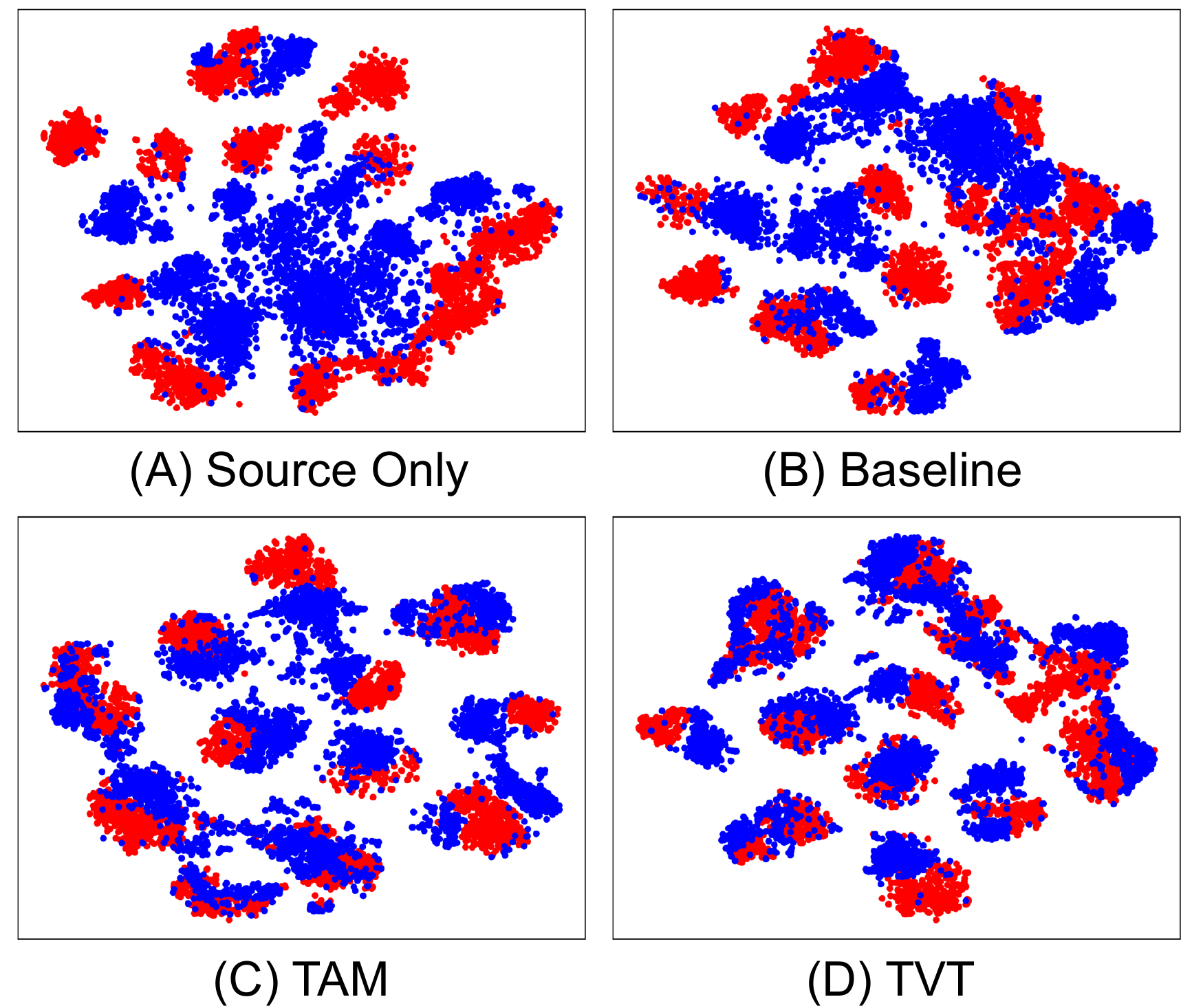}
	\end{center}
	\caption{t-SNE visualization of VisDA-2017 dataset, where red and blue points indicate the source (synthetic rendering) and the target (real images) domain, respectively.}
	\label{fig:tsne}
%	\vspace{-0.2in}
\end{figure}

For object recognition task, Office-31, Office-Home, and VisDA-2017 are used in evaluation.
As shown in Table~\ref{table:office31}~\ref{table:office_home},~\ref{table:visda}, TVT sets up new benchmark results for all the three datasets.
On the medium-sized Office-Home dataset (Table~\ref{table:office_home}), we achieve the significant improvement over the best prior UDA method (83.56\% vs 71.8\%).  
Results on the large-scale VisDA-2017 dataset (Table~\ref{table:visda}) show that we not only achieve a higher average accuracy, but also compete favorably against ALDA and SHOT that rely on pseudo labels. We believe training with pseudo label would give TVT extra accuracy gain, while it is out of our current scope.
Note that DTA also enforces the cluster assumption to learn discriminative features, but it fails to encourage the global diversity which may leads to a degenerate solution where every point is assigned to the same class.
% It is worth mentioning that ViT-based source only already outperforms previous state of the arts which are built upon AlexNet and ResNet, revealing the impressing transferability of ViT.
Besides, TVT surpasses both Source Only and Baseline, revealing its effectiveness in transferring domain knowledge by (i) capturing both transferable and discriminative fine-grained features and (ii) retaining discriminative information while searching for the domain-invariant representations.
This is also evidenced by the t-SNE visualization of learned features as showcased in Figure~\ref{fig:tsne}.
Obviously, TAM can effectively align source and target domain features by exploiting the local feature transferability.
However, the target feature is not well-separated due to that target labels in training are absent and the discriminative information are destroyed by adversarial alignment.
Fortunately, this problem is alleviated by DCM by assuming that datapoints should be classified with large margin, as illustrated in Figure~\ref{fig:tsne} (D).  

% Despite this, the intrinsic characteristics of ViT are still not fully explored for knowledge transfer as we have discussed.
% Our method, instead, proposes to use T-MSA to explicitly consider the transferability of local features during capturing long-range dependencies, giving rise to further improvement.
%In a nutshell, we find that robust representation learnt from adversarial pretraining is transferable to down-stream fine-tuning tasks to some extent.
%DTA (Drop to Adapt: Learning Discriminative Features for Unsupervised Domain Adaptation) is 81.5 on VisDA through learning strongly discriminative features by enforcing the cluster assumption
%The following methods are good at VisDA but they either based on pseudo labels or ensemble of multiple predictions:
%FixBi: Bridging Domain Spaces for Unsupervised Domain Adaptation
%Contrastive Adaptation Network for Unsupervised Domain Adaptation
%Self-ensembling for visual domain adaptation

%\vspace{-10pt}
\paragraph{Ablation Study}
To learn the individual contribution of TAM and DCM in improving the knowledge transferability of ViT, we conduct the ablation study in Table~\ref{table:ablation}.
Compared to Source Only, TAM consistently improves the classification accuracy with average 4.82\% boost, indicating the significance of capturing both transferable and discriminative features.
The performance is further improved by incorporating DCM, justifying the necessary of retaining the discriminative information of the learned representation. 
It is noteworthy that DCM brings the largest improvement on the large-scale synthetic-to-real VisDA-2017 dataset.
We suspect that the large domain gap in {VisDA-2017} (synthetic 2D rendering to natural image) is the leading reason, since simply aligning two domains with large domain shift results in a mess distributed feature space.
This challenge, however, can be largely addressed by DCM that enables retaining discriminative information based on a cluster assumption.

\begin{table}
	\footnotesize
	\setlength\tabcolsep{2pt}
	\begin{center}
		\begin{tabularx}{.47\textwidth}{ ccccc|g @{} }
			\toprule
			Methods & Digits & Office-31 & Office-Home & VisDA-2017 & Avg \\
			\midrule
			Source Only & 88.30 & 89.45 & 78.74 & 73.22 & 82.43 \\
			+TAM & 97.20 & 91.21 & 81.30 & 79.30 & 87.25 \\
			+DCM & 98.87 & 93.85 & 83.56 & 83.92 & 90.05 \\
			\bottomrule
		\end{tabularx}
	\end{center}
	\caption{Ablation study of each module.}
	\label{table:ablation}
	\vspace{-0.2in}
\end{table}

\begin{figure}[t]
	\begin{center}
		\includegraphics[width=1\linewidth]{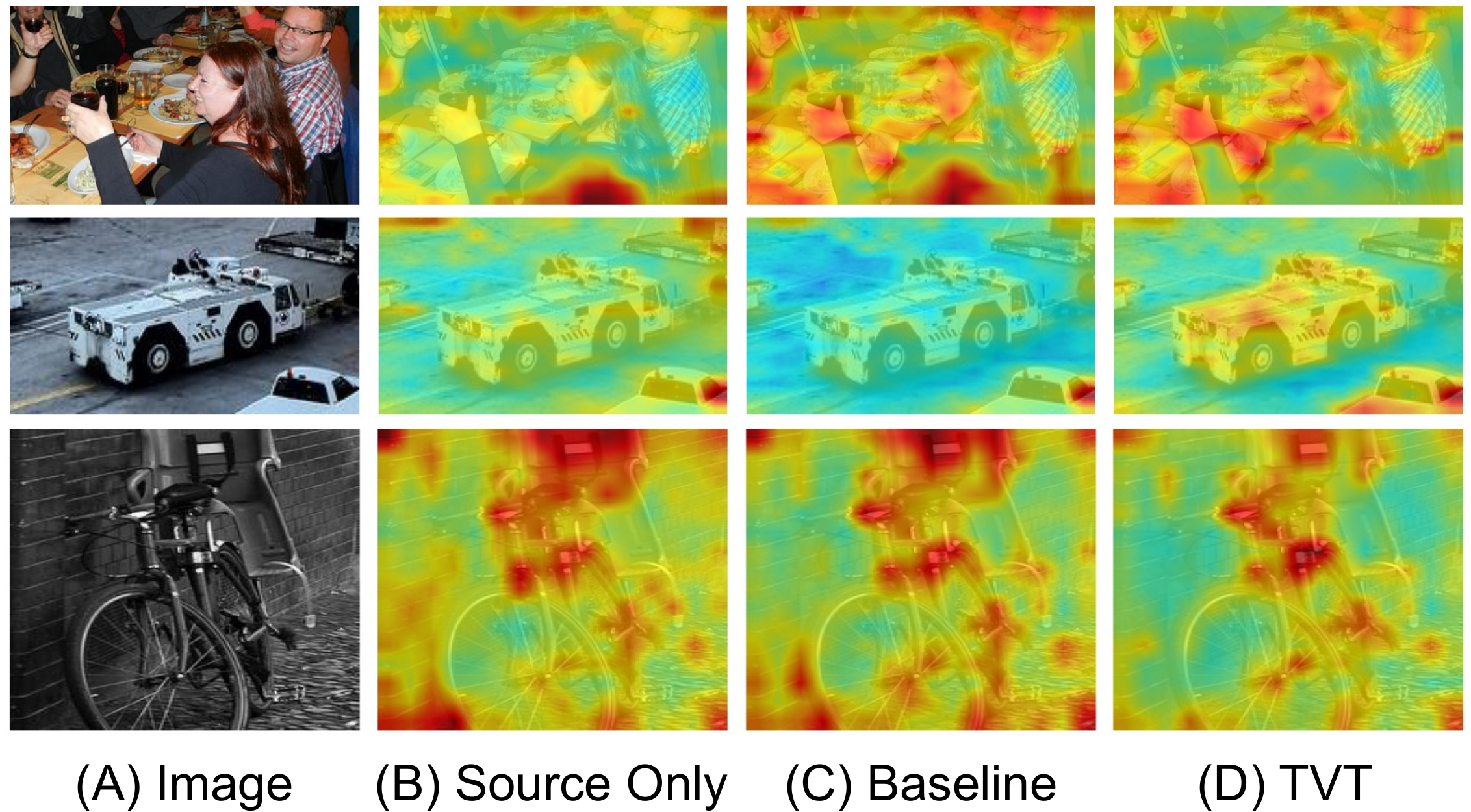}
	\end{center}
	\caption{Attention map visualization of person, truck, and bicycle in VisDA-2017 dataset. The hotter the color, the higher the attention.}
	\label{fig:attention}
	\vspace{-0.2in}
\end{figure}

%\vspace{-10pt}
\paragraph{Attention Visualization}
We visualize the attention map of the class token in TAM to verify that our model can attend to local features that are both transferable and discriminative.
Without loss of generality, we randomly sample target-domain images in VisDA-2017 dataset for comparison.
As shown in Figure~\ref{fig:attention}, our method captures more accurate regions than Source Only and Baseline.
For instance, to recognize the person in the top-left image, Source Only mainly focus on women's shoulder which is discriminative yet not highly transferable.
Moving beyond the shoulder region, the baseline also attends to faces and hands that can generalize well across domains.
Our method, instead, ignores the shoulder and only highlight those regions that are important for classification and transferable.
Certainly, by leveraging the intrinsic attention mechanism and fine-grained features captured by sequential patches, our method promotes the capability of ViT in transferring domain knowledge.

\vspace{-3pt}
\section{Conclusion}
In this paper, we perform the first-of-its-kind investigation of ViT's transferability in UDA task and observe that ViT are more transferable than CNNs counterparts.
To further improve the power of ViT in transferring domain knowledge, we propose TVT by explicitly considering the intrinsic merits of transformer architecture.
Specifically, TVT captures both transferable and discriminative features in the given image, and retains discriminative information of the learnt domain-invariant representations.
Experimental results on widely used benchmarks show that TVT outperforms prior UDA methods by a large margin.

{\bibliography{aaai22.bib}}

\begin{thebibliography}{51}
\providecommand{\natexlab}[1]{#1}

\bibitem[{Bello et~al.(2019)Bello, Zoph, Vaswani, Shlens, and
  Le}]{bello2019attention}
Bello, I.; Zoph, B.; Vaswani, A.; Shlens, J.; and Le, Q.~V. 2019.
\newblock Attention augmented convolutional networks.
\newblock In \emph{Proceedings of the IEEE/CVF International Conference on
  Computer Vision}, 3286--3295.

\bibitem[{Beltagy, Peters, and Cohan(2020)}]{beltagy2020longformer}
Beltagy, I.; Peters, M.~E.; and Cohan, A. 2020.
\newblock Longformer: The long-document transformer.
\newblock \emph{arXiv preprint arXiv:2004.05150}.

\bibitem[{Borgwardt et~al.(2006)Borgwardt, Gretton, Rasch, Kriegel,
  Sch{\"o}lkopf, and Smola}]{borgwardt2006integrating}
Borgwardt, K.~M.; Gretton, A.; Rasch, M.~J.; Kriegel, H.-P.; Sch{\"o}lkopf, B.;
  and Smola, A.~J. 2006.
\newblock Integrating structured biological data by kernel maximum mean
  discrepancy.
\newblock \emph{Bioinformatics}, 22(14): e49--e57.

\bibitem[{Bridle, Heading, and MacKay(1992)}]{bridle1992unsupervised}
Bridle, J.~S.; Heading, A.~J.; and MacKay, D.~J. 1992.
\newblock Unsupervised classifiers, mutual information and'phantom targets'.

\bibitem[{Carion et~al.(2020)Carion, Massa, Synnaeve, Usunier, Kirillov, and
  Zagoruyko}]{carion2020end}
Carion, N.; Massa, F.; Synnaeve, G.; Usunier, N.; Kirillov, A.; and Zagoruyko,
  S. 2020.
\newblock End-to-end object detection with transformers.
\newblock In \emph{European Conference on Computer Vision}, 213--229. Springer.

\bibitem[{Chapelle and Zien(2005)}]{chapelle2005semi}
Chapelle, O.; and Zien, A. 2005.
\newblock Semi-supervised classification by low density separation.
\newblock In \emph{International workshop on artificial intelligence and
  statistics}, 57--64. PMLR.

\bibitem[{Chen et~al.(2019)Chen, Xie, Huang, Rong, Ding, Huang, Xu, and
  Huang}]{chen2019progressive}
Chen, C.; Xie, W.; Huang, W.; Rong, Y.; Ding, X.; Huang, Y.; Xu, T.; and Huang,
  J. 2019.
\newblock Progressive feature alignment for unsupervised domain adaptation.
\newblock In \emph{Proceedings of the IEEE/CVF Conference on Computer Vision
  and Pattern Recognition}, 627--636.

\bibitem[{Chen et~al.(2020)Chen, Zhao, Liu, and Cai}]{chen2020adversarial}
Chen, M.; Zhao, S.; Liu, H.; and Cai, D. 2020.
\newblock Adversarial-learned loss for domain adaptation.
\newblock In \emph{Proceedings of the AAAI Conference on Artificial
  Intelligence}, volume~34, 3521--3528.

\bibitem[{Deng et~al.(2009)Deng, Dong, Socher, Li, Li, and
  Fei-Fei}]{deng2009imagenet}
Deng, J.; Dong, W.; Socher, R.; Li, L.-J.; Li, K.; and Fei-Fei, L. 2009.
\newblock Imagenet: A large-scale hierarchical image database.
\newblock In \emph{2009 IEEE conference on computer vision and pattern
  recognition}, 248--255. Ieee.

\bibitem[{Devlin et~al.(2018)Devlin, Chang, Lee, and
  Toutanova}]{devlin2018bert}
Devlin, J.; Chang, M.-W.; Lee, K.; and Toutanova, K. 2018.
\newblock Bert: Pre-training of deep bidirectional transformers for language
  understanding.
\newblock \emph{arXiv preprint arXiv:1810.04805}.

\bibitem[{Dosovitskiy et~al.(2020)Dosovitskiy, Beyer, Kolesnikov, Weissenborn,
  Zhai, Unterthiner, Dehghani, Minderer, Heigold, Gelly
  et~al.}]{dosovitskiy2020image}
Dosovitskiy, A.; Beyer, L.; Kolesnikov, A.; Weissenborn, D.; Zhai, X.;
  Unterthiner, T.; Dehghani, M.; Minderer, M.; Heigold, G.; Gelly, S.; et~al.
  2020.
\newblock An image is worth 16x16 words: Transformers for image recognition at
  scale.
\newblock \emph{arXiv preprint arXiv:2010.11929}.

\bibitem[{Ganin and Lempitsky(2015)}]{ganin2015unsupervised}
Ganin, Y.; and Lempitsky, V. 2015.
\newblock Unsupervised domain adaptation by backpropagation.
\newblock In \emph{International conference on machine learning}, 1180--1189.
  PMLR.

\bibitem[{Ganin et~al.(2016)Ganin, Ustinova, Ajakan, Germain, Larochelle,
  Laviolette, Marchand, and Lempitsky}]{ganin2016domain}
Ganin, Y.; Ustinova, E.; Ajakan, H.; Germain, P.; Larochelle, H.; Laviolette,
  F.; Marchand, M.; and Lempitsky, V. 2016.
\newblock Domain-adversarial training of neural networks.
\newblock \emph{The journal of machine learning research}, 17(1): 2096--2030.

\bibitem[{Girdhar et~al.(2019)Girdhar, Carreira, Doersch, and
  Zisserman}]{girdhar2019video}
Girdhar, R.; Carreira, J.; Doersch, C.; and Zisserman, A. 2019.
\newblock Video action transformer network.
\newblock In \emph{Proceedings of the IEEE/CVF Conference on Computer Vision
  and Pattern Recognition}, 244--253.

\bibitem[{Gomes, Krause, and Perona(2010)}]{gomes2010discriminative}
Gomes, R.; Krause, A.; and Perona, P. 2010.
\newblock Discriminative clustering by regularized information maximization.

\bibitem[{Goodfellow et~al.(2014)Goodfellow, Pouget-Abadie, Mirza, Xu,
  Warde-Farley, Ozair, Courville, and Bengio}]{goodfellow2014generative}
Goodfellow, I.~J.; Pouget-Abadie, J.; Mirza, M.; Xu, B.; Warde-Farley, D.;
  Ozair, S.; Courville, A.; and Bengio, Y. 2014.
\newblock Generative adversarial networks.
\newblock \emph{arXiv preprint arXiv:1406.2661}.

\bibitem[{He et~al.(2016)He, Zhang, Ren, and Sun}]{he2016deep}
He, K.; Zhang, X.; Ren, S.; and Sun, J. 2016.
\newblock Deep residual learning for image recognition.
\newblock In \emph{Proceedings of the IEEE conference on computer vision and
  pattern recognition}, 770--778.

\bibitem[{Hoffman et~al.(2018)Hoffman, Tzeng, Park, Zhu, Isola, Saenko, Efros,
  and Darrell}]{hoffman2018cycada}
Hoffman, J.; Tzeng, E.; Park, T.; Zhu, J.-Y.; Isola, P.; Saenko, K.; Efros, A.;
  and Darrell, T. 2018.
\newblock Cycada: Cycle-consistent adversarial domain adaptation.
\newblock In \emph{International conference on machine learning}, 1989--1998.
  PMLR.

\bibitem[{Hu et~al.(2018)Hu, Gu, Zhang, Dai, and Wei}]{hu2018relation}
Hu, H.; Gu, J.; Zhang, Z.; Dai, J.; and Wei, Y. 2018.
\newblock Relation networks for object detection.
\newblock In \emph{Proceedings of the IEEE Conference on Computer Vision and
  Pattern Recognition}, 3588--3597.

\bibitem[{Hu et~al.(2017)Hu, Miyato, Tokui, Matsumoto, and
  Sugiyama}]{hu2017learning}
Hu, W.; Miyato, T.; Tokui, S.; Matsumoto, E.; and Sugiyama, M. 2017.
\newblock Learning discrete representations via information maximizing
  self-augmented training.
\newblock In \emph{International conference on machine learning}, 1558--1567.
  PMLR.

\bibitem[{Krizhevsky, Sutskever, and Hinton(2012)}]{krizhevsky2012imagenet}
Krizhevsky, A.; Sutskever, I.; and Hinton, G.~E. 2012.
\newblock Imagenet classification with deep convolutional neural networks.
\newblock \emph{Advances in neural information processing systems}, 25:
  1097--1105.

\bibitem[{LeCun et~al.(1998)LeCun, Bottou, Bengio, and
  Haffner}]{lecun1998gradient}
LeCun, Y.; Bottou, L.; Bengio, Y.; and Haffner, P. 1998.
\newblock Gradient-based learning applied to document recognition.
\newblock \emph{Proceedings of the IEEE}, 86(11): 2278--2324.

\bibitem[{Lee et~al.(2019)Lee, Kim, Kim, and Jeong}]{lee2019drop}
Lee, S.; Kim, D.; Kim, N.; and Jeong, S.-G. 2019.
\newblock Drop to adapt: Learning discriminative features for unsupervised
  domain adaptation.
\newblock In \emph{Proceedings of the IEEE/CVF International Conference on
  Computer Vision}, 91--100.

\bibitem[{Liang, Hu, and Feng(2020)}]{liang2020we}
Liang, J.; Hu, D.; and Feng, J. 2020.
\newblock Do we really need to access the source data? source hypothesis
  transfer for unsupervised domain adaptation.
\newblock In \emph{International Conference on Machine Learning}, 6028--6039.
  PMLR.

\bibitem[{Liu et~al.(2019)Liu, Long, Wang, and Jordan}]{liu2019transferable}
Liu, H.; Long, M.; Wang, J.; and Jordan, M. 2019.
\newblock Transferable adversarial training: A general approach to adapting
  deep classifiers.
\newblock In \emph{International Conference on Machine Learning}, 4013--4022.
  PMLR.

\bibitem[{Liu et~al.(2021)Liu, Lin, Cao, Hu, Wei, Zhang, Lin, and
  Guo}]{liu2021swin}
Liu, Z.; Lin, Y.; Cao, Y.; Hu, H.; Wei, Y.; Zhang, Z.; Lin, S.; and Guo, B.
  2021.
\newblock Swin transformer: Hierarchical vision transformer using shifted
  windows.
\newblock \emph{arXiv preprint arXiv:2103.14030}.

\bibitem[{Long et~al.(2015)Long, Cao, Wang, and Jordan}]{long2015learning}
Long, M.; Cao, Y.; Wang, J.; and Jordan, M. 2015.
\newblock Learning transferable features with deep adaptation networks.
\newblock In \emph{International conference on machine learning}, 97--105.
  PMLR.

\bibitem[{Long et~al.(2017{\natexlab{a}})Long, Cao, Wang, and
  Jordan}]{long2017conditional}
Long, M.; Cao, Z.; Wang, J.; and Jordan, M.~I. 2017{\natexlab{a}}.
\newblock Conditional adversarial domain adaptation.
\newblock \emph{arXiv preprint arXiv:1705.10667}.

\bibitem[{Long et~al.(2017{\natexlab{b}})Long, Zhu, Wang, and
  Jordan}]{long2017deep}
Long, M.; Zhu, H.; Wang, J.; and Jordan, M.~I. 2017{\natexlab{b}}.
\newblock Deep transfer learning with joint adaptation networks.
\newblock In \emph{International conference on machine learning}, 2208--2217.
  PMLR.

\bibitem[{Neimark et~al.(2021)Neimark, Bar, Zohar, and
  Asselmann}]{neimark2021video}
Neimark, D.; Bar, O.; Zohar, M.; and Asselmann, D. 2021.
\newblock Video transformer network.
\newblock \emph{arXiv preprint arXiv:2102.00719}.

\bibitem[{Netzer et~al.(2011)Netzer, Wang, Coates, Bissacco, Wu, and
  Ng}]{netzer2011reading}
Netzer, Y.; Wang, T.; Coates, A.; Bissacco, A.; Wu, B.; and Ng, A.~Y. 2011.
\newblock Reading digits in natural images with unsupervised feature learning.

\bibitem[{Pan and Yang(2009)}]{pan2009survey}
Pan, S.~J.; and Yang, Q. 2009.
\newblock A survey on transfer learning.
\newblock \emph{IEEE Transactions on knowledge and data engineering}, 22(10):
  1345--1359.

\bibitem[{Pei et~al.(2018)Pei, Cao, Long, and Wang}]{pei2018multi}
Pei, Z.; Cao, Z.; Long, M.; and Wang, J. 2018.
\newblock Multi-adversarial domain adaptation.
\newblock In \emph{Proceedings of the AAAI Conference on Artificial
  Intelligence}, volume~32.

\bibitem[{Peng et~al.(2017)Peng, Usman, Kaushik, Hoffman, Wang, and
  Saenko}]{peng2017visda}
Peng, X.; Usman, B.; Kaushik, N.; Hoffman, J.; Wang, D.; and Saenko, K. 2017.
\newblock Visda: The visual domain adaptation challenge.
\newblock \emph{arXiv preprint arXiv:1710.06924}.

\bibitem[{Saenko et~al.(2010)Saenko, Kulis, Fritz, and
  Darrell}]{saenko2010adapting}
Saenko, K.; Kulis, B.; Fritz, M.; and Darrell, T. 2010.
\newblock Adapting visual category models to new domains.
\newblock In \emph{European conference on computer vision}, 213--226. Springer.

\bibitem[{Saito et~al.(2018)Saito, Watanabe, Ushiku, and
  Harada}]{saito2018maximum}
Saito, K.; Watanabe, K.; Ushiku, Y.; and Harada, T. 2018.
\newblock Maximum classifier discrepancy for unsupervised domain adaptation.
\newblock In \emph{Proceedings of the IEEE conference on computer vision and
  pattern recognition}, 3723--3732.

\bibitem[{Shi and Sha(2012)}]{shi2012information}
Shi, Y.; and Sha, F. 2012.
\newblock Information-theoretical learning of discriminative clusters for
  unsupervised domain adaptation.
\newblock \emph{arXiv preprint arXiv:1206.6438}.

\bibitem[{Tzeng et~al.(2017)Tzeng, Hoffman, Saenko, and
  Darrell}]{tzeng2017adversarial}
Tzeng, E.; Hoffman, J.; Saenko, K.; and Darrell, T. 2017.
\newblock Adversarial discriminative domain adaptation.
\newblock In \emph{Proceedings of the IEEE conference on computer vision and
  pattern recognition}, 7167--7176.

\bibitem[{Tzeng et~al.(2014)Tzeng, Hoffman, Zhang, Saenko, and
  Darrell}]{tzeng2014deep}
Tzeng, E.; Hoffman, J.; Zhang, N.; Saenko, K.; and Darrell, T. 2014.
\newblock Deep domain confusion: maximizing for domain invariance (2014).
\newblock \emph{Preprint. arXiv}, 1412.

\bibitem[{Vaswani et~al.(2017)Vaswani, Shazeer, Parmar, Uszkoreit, Jones,
  Gomez, Kaiser, and Polosukhin}]{vaswani2017attention}
Vaswani, A.; Shazeer, N.; Parmar, N.; Uszkoreit, J.; Jones, L.; Gomez, A.~N.;
  Kaiser, L.; and Polosukhin, I. 2017.
\newblock Attention is all you need.
\newblock \emph{arXiv preprint arXiv:1706.03762}.

\bibitem[{Venkateswara et~al.(2017)Venkateswara, Eusebio, Chakraborty, and
  Panchanathan}]{venkateswara2017deep}
Venkateswara, H.; Eusebio, J.; Chakraborty, S.; and Panchanathan, S. 2017.
\newblock Deep hashing network for unsupervised domain adaptation.
\newblock In \emph{Proceedings of the IEEE conference on computer vision and
  pattern recognition}, 5018--5027.

\bibitem[{Wang and Deng(2018)}]{wang2018deep}
Wang, M.; and Deng, W. 2018.
\newblock Deep visual domain adaptation: A survey.
\newblock \emph{Neurocomputing}, 312: 135--153.

\bibitem[{Wang et~al.(2021)Wang, Xie, Li, Fan, Song, Liang, Lu, Luo, and
  Shao}]{wang2021pyramid}
Wang, W.; Xie, E.; Li, X.; Fan, D.-P.; Song, K.; Liang, D.; Lu, T.; Luo, P.;
  and Shao, L. 2021.
\newblock Pyramid vision transformer: A versatile backbone for dense prediction
  without convolutions.
\newblock \emph{arXiv preprint arXiv:2102.12122}.

\bibitem[{Wang et~al.(2018)Wang, Girshick, Gupta, and He}]{wang2018non}
Wang, X.; Girshick, R.; Gupta, A.; and He, K. 2018.
\newblock Non-local neural networks.
\newblock In \emph{Proceedings of the IEEE conference on computer vision and
  pattern recognition}, 7794--7803.

\bibitem[{Wang et~al.(2019)Wang, Li, Ye, Long, and Wang}]{wang2019transferable}
Wang, X.; Li, L.; Ye, W.; Long, M.; and Wang, J. 2019.
\newblock Transferable attention for domain adaptation.
\newblock In \emph{Proceedings of the AAAI Conference on Artificial
  Intelligence}, volume~33, 5345--5352.

\bibitem[{Wang et~al.(2020)Wang, Xu, Wang, Shen, Cheng, Shen, and
  Xia}]{wang2020end}
Wang, Y.; Xu, Z.; Wang, X.; Shen, C.; Cheng, B.; Shen, H.; and Xia, H. 2020.
\newblock End-to-End Video Instance Segmentation with Transformers.
\newblock \emph{arXiv preprint arXiv:2011.14503}.

\bibitem[{Ying et~al.(2018)Ying, Zhang, Huang, and Yang}]{ying2018transfer}
Ying, W.; Zhang, Y.; Huang, J.; and Yang, Q. 2018.
\newblock Transfer learning via learning to transfer.
\newblock In \emph{International Conference on Machine Learning}, 5085--5094.
  PMLR.

\bibitem[{Yosinski et~al.(2014)Yosinski, Clune, Bengio, and
  Lipson}]{yosinski2014transferable}
Yosinski, J.; Clune, J.; Bengio, Y.; and Lipson, H. 2014.
\newblock How transferable are features in deep neural networks?
\newblock \emph{arXiv preprint arXiv:1411.1792}.

\bibitem[{Zheng et~al.(2020)Zheng, Lu, Zhao, Zhu, Luo, Wang, Fu, Feng, Xiang,
  Torr et~al.}]{zheng2020rethinking}
Zheng, S.; Lu, J.; Zhao, H.; Zhu, X.; Luo, Z.; Wang, Y.; Fu, Y.; Feng, J.;
  Xiang, T.; Torr, P.~H.; et~al. 2020.
\newblock Rethinking Semantic Segmentation from a Sequence-to-Sequence
  Perspective with Transformers.
\newblock \emph{arXiv preprint arXiv:2012.15840}.

\bibitem[{Zhou et~al.(2020)Zhou, Zhang, Peng, Zhang, Li, Xiong, and
  Zhang}]{zhou2020informer}
Zhou, H.; Zhang, S.; Peng, J.; Zhang, S.; Li, J.; Xiong, H.; and Zhang, W.
  2020.
\newblock Informer: Beyond Efficient Transformer for Long Sequence Time-Series
  Forecasting.
\newblock \emph{arXiv preprint arXiv:2012.07436}.

\bibitem[{Zhu et~al.(2020)Zhu, Su, Lu, Li, Wang, and Dai}]{zhu2020deformable}
Zhu, X.; Su, W.; Lu, L.; Li, B.; Wang, X.; and Dai, J. 2020.
\newblock Deformable detr: Deformable transformers for end-to-end object
  detection.
\newblock \emph{arXiv preprint arXiv:2010.04159}.

\end{thebibliography}
\bibliographystyle{aaai}

\end{document}